\documentclass{article}
\PassOptionsToPackage{numbers, compress, sort}{natbib}

\usepackage[preprint]{neurips_2021}
\usepackage[utf8]{inputenc} 
\usepackage[T1]{fontenc}    
\usepackage{hyperref}       
\usepackage{url}            
\usepackage{booktabs}       
\usepackage{amsfonts,amsmath, amssymb}       
\usepackage{nicefrac}       
\usepackage{microtype}      

\usepackage{lipsum}
\usepackage{nicefrac}       
\usepackage{float}

\usepackage{amsmath,amsthm}
\usepackage{bm}
\usepackage{mathtools}
\usepackage{xcolor}
\definecolor{grey}{rgb}{0.33, 0.33, 0.33}
\usepackage{wrapfig}
\usepackage{stmaryrd}
\usepackage{tikz}
\usepackage{pgfplots}
\pgfplotsset{compat=1.14}

\usepackage[title]{appendix}

\newcommand{\titl}{Boosting Graph Neural Networks by Injecting Pooling in Message Passing}
\usepackage{caption}
\usepackage{color}

\makeatletter

\newcommand\Autoref[1]{\@first@ref#1,@}
\def\@throw@dot#1.#2@{#1}
\def\@set@refname#1{
    \edef\@tmp{\getrefbykeydefault{#1}{anchor}{}}%
    \xdef\@tmp{\expandafter\@throw@dot\@tmp.@}%
    \ltx@IfUndefined{\@tmp autorefnameplural}%
         {\def\@refname{\@nameuse{\@tmp autorefname}s}}%
         {\def\@refname{\@nameuse{\@tmp autorefnameplural}}}%
}
\def\@first@ref#1,#2{%
  \ifx#2@\autoref{#1}\let\@nextref\@gobble
  \else%
    \@set@refname{#1}
    \@refname~\ref{#1}
    \let\@nextref\@next@ref
  \fi%
  \@nextref#2%
}
\def\@next@ref#1,#2{%
   \ifx#2@ and~\ref{#1}\let\@nextref\@gobble
   \else, \ref{#1}
   \fi%
   \@nextref#2%
}

\makeatother

\title{\titl}

\newcommand{\authorinfo}{
 Hyeokjin Kwon, Jong-Min Lee \\
 Hanyang University, Republic of Korea\\
  \texttt{\{kweon186, ljm\}@hanyang.ac.kr} 
}

\author{\authorinfo}

\usepackage{xcolor,colortbl}
\definecolor{Gray}{gray}{0.85}

\newcommand{\transpose}{\intercal}
\usepackage{mathtools}

\usepackage{bchart}

\definecolor{blue}{RGB}{31,119,180}
\definecolor{orange}{RGB}{255,127,14}
\definecolor{red}{RGB}{214,39,40}
\definecolor{green}{RGB}{44,160,44}

\begin{document}
\maketitle

\begin{abstract}
There has been tremendous success in the field of graph neural networks (GNNs) as a result of the development of the message-passing (MP) layer, which updates the representation of a node by combining it with its neighbors to address variable-size and unordered graphs. Despite the fruitful progress of MP GNNs, their performance can suffer from over-smoothing, when node representations become too similar and even indistinguishable from one another. Furthermore, it has been reported that intrinsic graph structures are smoothed out as the GNN layer increases. Inspired by the edge-preserving bilateral filters used in image processing, we propose a new, adaptable, and powerful MP framework to prevent over-smoothing. Our \emph{bilateral}-MP estimates a pairwise modular gradient by utilizing the class information of nodes, and further preserves the global graph structure by using the gradient when the aggregating function is applied. Our proposed scheme can be generalized to all ordinary MP GNNs. Experiments on five medium-size benchmark datasets using four state-of-the-art MP GNNs indicate that the \emph{bilateral}-MP improves performance by alleviating over-smoothing. By inspecting quantitative measurements, we additionally validate the effectiveness of the proposed mechanism in preventing the over-smoothing issue.
\end{abstract}
\section{Introduction}
\label{sec:intro}
Unlike regular data structures such as images and language sequences, graphs are frequently used to model data with arbitrary topology in many fields of science and engineering~\cite{agcn, reviewGNN, asap}. Graph neural networks (GNNs) are promising methods that are used for major graph representation tasks including graph classification, edge prediction, and node classification~\cite{pna, gmm, klicpera2019predict}. In recent years, the most widely used methods for GNNs have learnt node representations by gradually aggregating local neighbors. This is known as message-passing (MP)~\cite{dwivedi2020benchmarking, sukhbaatar2016learning, kipf2017semisupervised, hamilton2018inductive}. 

Although MP GNNs have proven to be successful in various fields, it has been reported that MP makes node representations closer, leading them to inevitably converge to indistinguishable values with the increase in network depth~\cite{zhao2020pairnorm, dgn}. This issue, called over-smoothing, impedes the MP GNNs by making all nodes unrelated to the input features and further hurts prediction performance by preventing the model from going deeper~\cite{li2019deepgcns, klicpera2019predict}. The main cause of over-smoothing is the over-mixing of helpful interaction messages and noise from other node pairs belonging to different classes~\cite{dgn, chen2019measuring}. Recently, several methods have been developed to alleviate over-smoothing through various normalization schemes, including batch~\cite{ioffe2015batch, dwivedi2020benchmarking}, pair~\cite{zhao2020pairnorm}, and group normalization~\cite{dgn}, as well as message reduction via dropping a subset of edges~\cite{rong2020dropedge} or nodes~\cite{chen2019measuring}. However, these methods do not directly consider the fundamental problems of MP GNNs.

\begin{figure}[t!]
    \centering
    \includegraphics[width=\textwidth]{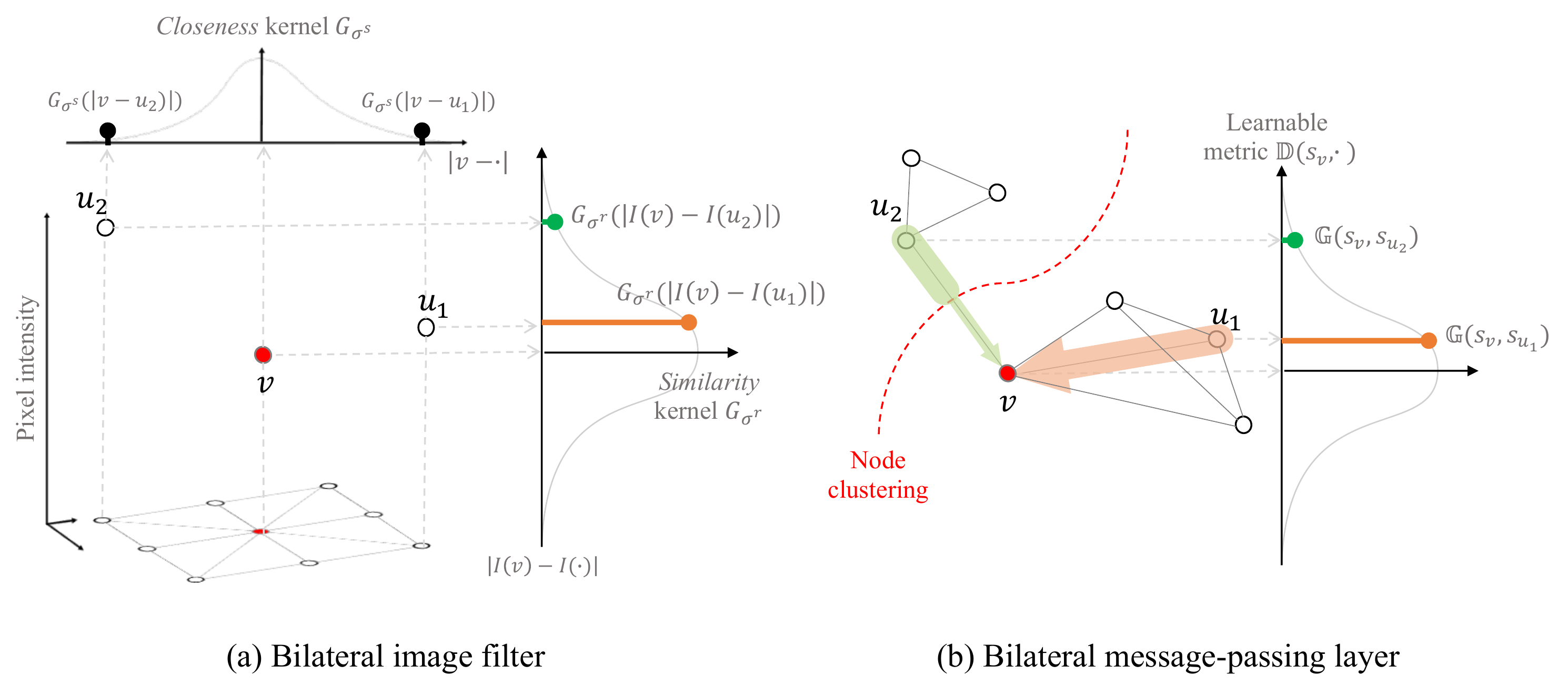}
    \caption{Comparison between (a) the bilateral image filter and (b) proposed \emph{bilateral}-MP layer. The white and red circles denote a set of neighborhood nodes ($u \in \mathcal{N}_{v}$) and a center node ($v$), respectively. The bilateral image filter smooths the target image while preserving the edge structure by utilizing \textcolor{orange}{the gradient of intensity information $I(\cdot)$} of adjacent pixels (nodes): $I_{bi}(v)={\frac 1 N} \sum_{u \in V}\mathbb{G}_{\sigma^s}(|v-u|)\textcolor{orange}{\mathbb{G}_{\sigma^r}(|I(v)-I(u)|)}I(v)$ where $\mathbb{G}_{\sigma^s}$ is conventional filter kernel of \emph{closeness}, and $\mathbb{G}_{\sigma^r}$ is regulating kernel of \emph{similarity}. Similarly, given the soft class information $\bm{s}_{v}$ for the node $v$, the proposed \emph{bilateral}-MP layer calculates the modular gradient $\mathbb{G}(\bm{s}_{v},\bm{s}_{u}), \forall u \in \mathcal{N}_{v}$ and regulates the aggregating functions to prevent over-smoothing.}
    \label{fig:Figure1}
\end{figure}

In the context of computer vision tasks, some studies may provide insights into this issue. Intuitively, the MP operation can be considered analogous to image filtering, which uses a spatial kernel to extract the weighted sum of values from local neighbors (e.g., mean filter and median filter). The concept underlying image filtering is slow spatial variation, as adjacent pixels are likely to have similar values~\cite{tomasi1998bilateral}. However, this assumption clearly fails at the edges of objects, which serve as boundaries between different homogeneous regions~\cite{eps}. This poses a challenge for image filtering to preserve local structures. The bilateral filter is a common method that regulates filtering by enforcing not only \emph{closeness} but also \emph{similarity} to prevent over-mixing across edges~\cite{tomasi1998bilateral}. The left part of \autoref{fig:Figure1} illustrates an example of the bilateral filter. Inspired by this, one may expect a structure-preserving MP mechanism that regulates message aggregation using global information, such as the community structure of the nodes suitable for graphs, to prevent over-smoothing.

Analogous to the bilateral image filter, we propose a new, adaptable, and powerful MP scheme to address over-smoothing in classic MP GNNs. We designed a type of \emph{anisotropic} MP that leverages node classes to estimate the pairwise modular gradient, and further regulates message propagation from neighborhoods using the gradient for better node representation (The right part of \autoref{fig:Figure1}). Our experiments demonstrate superior performance of the proposed MP scheme over that of most existing diffused GNNs for tasks with the five medium-scale benchmark datasets. To further understand the effect of the proposed \emph{bilateral}-MP scheme, we analyze how the quantitative measurements of over-smoothing change as the number of layers increases. Our work is open-sourced via GitHub, and the code for all models, details for the benchmarking framework, and hyperparameters are available with the PyTorch~\cite{pytorch}, and DGL~\cite{dgl} frameworks: \textcolor{blue}{\texttt{https://github.com/hookhy/bi-MP}}.
\section{Method}
\label{sec:method}
In this study, we introduce a new \emph{bilateral}-MP scheme that can be applied to ordinary MP GNN layers to prevent over-smoothing. Instead of directly propagating information through local edges, the proposed model defines a pairwise modular gradient between nodes and uses it to apply a gating mechanism to the MP layer's aggregating function. More specifically, the \emph{bilateral}-MP takes a soft assignment matrix of as input and extracts the modular gradient by applying metric learning layers to selectively transfer the messages. The key intuition is that the propagation of useful information within the same node class survives while the extraneous noise between different classes is reduced. Thus, the \emph{bilateral}-MP layer results in better graph representation and improved performance by preventing over-smoothing.

\subsection{GNNs based on the message-passing }
\label{subsec:mpgnns}
Given a graph $G = (V, E)$ with set of nodes $V$ $(|V| = N)$, and edges $E$, we set the input node feature vectors $\bm{x}_v \in \mathbb{R}^{D \times 1}$ for node $v$,  where $\bm{X} = [\bm{x}_{1}, \dots ,\bm{x}_{N}]^\transpose$ denotes the feature matrix. Generally, MP GNNs consist of stacked MP layers that learn node representations by applying an update function after aggregating neighboring messages. The update function at layer $l$ is as follows:

\begin{equation}
    \bm{h}^{l+1}_{v} =  \sigma(\bm{U}^{l}_{1}\bm{h}^{l}_{v} + \sum_{u \in \mathcal{N}_{v}} \bm{w}_{u,v} \bm{U}^{l}_{2} \bm{h}^{l}_{u} )
    \label{eq:gnn_orig}
\end{equation}

where $\mathcal{N}_{v}$ is a set of neighboring nodes, $\bm{h}^{l}_{v}$ is the feature vector of the node $v$ at layer $l$, $\bm{h}^{0}_{v}$ is set to the input node attribute $\bm{x}_v$, and $\bm{U}^{l}_{1} \in \mathbb{R}^{D \times D}$, and $\bm{U}^{l}_{2} \in \mathbb{R}^{D \times D}$ are trainable parameters at layer $l$. $\bm{w}_{u,v}$ is the weighting factor between nodes $i$ and $j$, and can be scalar or vector (if vector, weighting operation should be performed by applying element-wise multiplication), and $\sigma$ is the nonlinearity. In the case of \emph{isotropic} MP GNNs, such as Graph Convolutional Networks (GCN)~\cite{kipf2017semisupervised} and GraphSAGE~\cite{hamilton2018inductive}, $\bm{w}_{u,v}$ can be treated as a scalar with a value of 1.

GNNs perform downstream tasks, including node classification, edge prediction, and graph-level prediction, by using additional task-specific layers. First, the task of node classification involves predicting the node label $\bm{y}_{v}$  of the input graph by utilizing the following linear layer:

\begin{equation}
    \bm{y}_{v} = \sigma(\bm{W}_\text{FC} \bm{h}^{L}_{v})
    \label{eq:nodecls}
\end{equation}

where $\bm{W}_\text{FC} \in \mathbb{R}^{C \times D}$ is a trainable parameter with $C$ classes of node labels, and $\bm{h}^{L}_{v}$ is the final node representation of the last MP layer. To predict the edge attributes $\bm{y}_{e}$, we can apply a linear layer to the concatenation of the source’s representations and determine the edge. We can also perform graph-level prediction by making the linear layer receive the summary vector $\bm{z}_{G}$ for the input graph $G$. This summary vector $\bm{z}_{G}$ can be obtained by applying the readout-layer, which uses arithmetic operators such as summation and averaging~\cite{cangea2018towards}.

\begin{align}
    \bm{y}_{e} = \sigma(\bm{W}_\text{FC} \text{Concat}(\bm{h}^{L}_{src},\bm{h}^{L}_{dst}) )
    \text{, and } \bm{y}_{G} = \sigma(\bm{W}_\text{FC} \bm{z}_{G})
    \label{eq:edge_and_graph}
\end{align}

In this study, we use the mean readout function for the output representation as follows: 

\begin{equation}
    \bm{z}_{G} = {\frac 1 N} \sum_{v \in V} \bm{h}^{L}_{v}
    \label{eq:readout}
\end{equation}

\subsection{Modular gradient}
\label{subsec:modulargradient}
To circumvent the over-smoothing issue, we intend to reduce the number of incoming messages from the nodes of different classes. This goal suggests the need for the simultaneous definition of 1) node class assignment and 2) measurement of the pairwise difference based on the class information. 

By following the node clustering strategy introduced in \cite{dgn}, we train a stack of linear layers using both supervised task-specific loss $\mathcal{L}_s$ and unsupervised loss $\mathcal{L}_u$  based on the minimum cut objective to obtain the node-level class assignment matrix~\cite{bianchi2019mincut}:

\begin{equation}
    \bm{S}^{l} = \text{softmax}( \text{ReLU} ( \bm{H}^{l} \bm{W}^{l}_{1}) \bm{W}^{l}_{2} )
    \label{eq:softassign}
\end{equation}
\begin{equation}
    \mathcal{L}_u = {\frac 1 L} \sum_l \mathcal{L}_{u}^{l}
    \label{eq:unsup_loss}
\end{equation}

where $\bm{W}^{l}_{1} \in \mathbb{R}^{D \times D}$, and $\bm{W}^{l}_{2} \in \mathbb{R}^{D \times K^l}$ are trainable parameters, and $\bm{H}^l =[\bm{h}_{1}^{l}, \dots ,\bm{h}_{N}^{l}]^\transpose$ is the matrix of the hidden representations. For each layer $l$, the unsupervised loss $\mathcal{L}_{u}^{l}$ consists of a spectral clustering loss $\mathcal{L}_{c}^{l}$ based on the k-normalized minimum cut, and an orthogonal loss $\mathcal{L}_{o}^{l}$ enabling unique assignment in terms of the node: 

\begin{equation}
   \mathcal{L}_{u}^{l} = \mathcal{L}_{c}^{l} + \mathcal{L}_{o}^{l} = -{\frac {\text{Tr}((\bm{S}^{l})^\transpose \tilde{\bm{A}} \bm{S}^{l})} {\text{Tr}((\bm{S}^{l})^\transpose \tilde{\bm{D}} \bm{S}^{l})} } + 
   \|{\frac {(\bm{S}^{l})^\transpose \bm{S}^{l}} {\|(\bm{S}^{l})^\transpose \bm{S}^{l} \|_{\bm{F}}}} 
   - {\frac {\bm{I}_{K}} {\sqrt{K}}}\|_{\bm{F}}
   \label{eq:mincut_loss}
\end{equation}

where $\text{Tr} (\cdot)$ is a trace operator and $\|\cdot\|_{\bm{F}}$ is the Frobenius norm of the matrix. The spectral clustering loss $\mathcal{L}_{c}^{l}$ ensures that connected nodes are grouped into the same cluster while partitioning the nodes in ${K}^{l}$ disjoint subsets. Minimizing the orthogonal loss $\mathcal{L}_{o}^{l}$ enables the cluster assignment vectors $\bm{s}_{v}^{l} \in \mathbb{R}^{K^l \times 1}, \forall  v=1, \dots, N$ to be orthogonal.

To fulfill the second goal, we propose a novel metric called the modular gradient, which quantifies the pairwise difference of class assignment among the nodes. Specifically, we employ the metric learning mechanism to estimate the generalized Mahalanobis distance $\beta_{u,v}^{l}$  between the soft cluster assign vectors for the pairs of nodes $u$, and $v$ as follows~\cite{agcn}:

\begin{equation}
   \beta_{u,v}^{l} = \text{exp}( {\frac {-\mathbb{D}(\bm{s}_{u}^{l},\bm{s}_{v}^{l})} {2{\sigma_{m}}^2}} ) 
   \label{eq:modular_grad}
\end{equation}

where $\text{exp} (\cdot)$ denotes an exponential. The sensitivity to the class information of the message passing can be controlled by the hyperparameter $\sigma_{m}$. The Mahalanobis distance kernel $\mathbb{D} (\cdot)$ is defined as:

\begin{equation}
   \mathbb{D}(\bm{s}_{u}^{l},\bm{s}_{v}^{l}) =  \sqrt{ (\bm{s}_{u}^{l} - \bm{s}_{v}^{l})^\transpose{\bm{M}^{l}}(\bm{s}_{u}^{l} - \bm{s}_{v}^{l}) }\text{, where } \bm{M}^{l} = \bm{W}_{m}^{l} {\bm{W}_{m}^{l}}^\transpose
   \label{eq:metric_learn}
\end{equation}

where $\bm{W}_{m}^{l} \in \mathbb{R}^{D \times D}$ is a trainable parameter and $\mathbb{D} (\cdot)$ is equal to the Euclidean distance metric when $\bm{M}^{l} = \bm{I}$. The modular gradient $\beta_{u,v}^{l}$ will have a large value when nodes $u$ and $v$ belong to the same class, and the opposite case occurs when their assignments are different.

\subsection{Bilateral message-passing}
\label{subsec:bilateralmp}
The computed modular gradients were further normalized across all neighbors using following function:

\begin{equation}
   \hat{\beta}_{u,v}^{l} = {\frac {\text{sigmoid}(\beta_{u,v}^{l})} {\sum_{\Dot{u} \in \mathcal{N}_{v}} {\text{sigmoid}(\beta_{\Dot{u},v}^{l})}}}
   \label{eq:norm}
\end{equation}

where $\text{sigmoid}(\cdot)$ is a sigmoid function. Subsequently, we employed a new MP scheme, the \emph{bilateral}-MP, by applying an additional gating mechanism using node class information to the MP layers of \autoref{eq:gnn_orig}.

\begin{equation}
    \bm{h}^{l+1}_{v} =  \sigma(\bm{U}^{l}_{1}\bm{h}^{l}_{v} + \sum_{u \in \mathcal{N}_{v}} \hat{\beta}_{u,v}^{l} \bm{w}_{u,v} \bm{U}^{l}_{2} \bm{h}^{l}_{u} )
    \label{eq:bi_gnn}
\end{equation}

All existing MP GNNs can be improved with the \emph{bilateral}-MP layer(s) by modulating the original layer with a modular gradient as they address the overmixing of harmful noise between different classes.
\section{Experiments}
\label{sec:experiments}
In this section, we perform three fundamental supervised graph learning tasks (graph prediction, node classification, and edge prediction) using five medium-size benchmark datasets from diverse domains to evaluate the effectiveness and robustness of the proposed framework. With a unified and reproducible framework of \cite{dwivedi2020benchmarking}, we cover four state-of-the-art (SOTA) MP GNNs: GCN, GraphSAGE, Graph Attention Networks (GAT)~\cite{gat}, and Gated GCN~\cite{bresson2017residual}. We also use the quantitative measurement for over-smoothing based on information theory to investigate how the measurement changes by applying the proposed framework.

\begin{table}
    \centering
    \caption{Statistics on five medium-size benchmark datasets. Columns 4 and 5 denote mean statistics for the number of nodes and edges, respectively. Columns 6 and 7 summarize the node and edge features (and dimension), respectively.}
    \resizebox{\textwidth}{!}{%
    \begin{tabular}{llcrrcc}
    \toprule
    \textbf{Task} & \textbf{Dataset} & \textbf{\#Graphs} & \textbf{\#Nodes} & \textbf{\#Edges} &  \textbf{Node feat (dim)} & \textbf{Edge feat (dim)} \\
    \midrule
    Graph regression  & ZINC & 12k & 23.16  & 49.83 & Atom type (28) & Bond type (4)    \\
    Graph classification  & SMNIST & 70k & 70.57  & 564.53 & intensity+Coordinates (3) & Distance (1)    \\
    Graph classification  & CIFAR10 & 60k & 117.63  & 941.07 & RGB+Coordinates (5) & Distance (1)    \\
    Edge prediction  & TSP & 12k & 275.76  & 6894.04 & Coordinates (2) & Distance (1)    \\
    Node classification  & CLUSTER & 12k & 117.20  & 4301.72 & Attributes (7) & $-$    \\ 
    \bottomrule
    \end{tabular}%
    }
    \label{tab:dataset}
\end{table}

\subsection{Dataset}
\label{subsec:dataset}
Following \cite{dwivedi2020benchmarking}, we used two artificially generated (TSP~\cite{joshi2019efficient}, and CLUSTER~\cite{sbm}) and three real-world datasets (SMINIST~\cite{lecun1998gradient}, CIFAR10~\cite{krizhevsky2009learning}, and ZINC~\cite{irwin2012zinc}) with various sizes and complexities, covering all graph tasks. For the graph regression task, we predicted the constrained solubility of the molecular graphs that were randomly selected from the ZINC dataset. We also used the super-pixel datasets SMNIST and CIFAR10, which were converted into a graph classification task using the super-pixels algorithm~\cite{achanta2012slic}. The Traveling Salesman problem, which the TSP dataset is based in, is one of the most intensively studied problems relating to the multiscale and complex nature of 2D Euclidean graphs. As shown in \cite{dwivedi2020benchmarking}, we cast the TSP as a binary edge prediction task that identifies the optimal edges given by the Concorde. CLUSTER, a semi-supervised node classification task, was generated using stochastic block models (SBM), which can be used to model community structures~\cite{sbm}. More details for the datasets are presented in \cite{dwivedi2020benchmarking}, and corresponding statistics are reported in \autoref{tab:dataset}.

\subsection{Benchmarking framework}
\label{subsec:bencharking}
For all datasets, we followed the same benchmarking protocol, which encompasses the method for data splits, pre-processing, performance metrics, and other training details in \cite{dwivedi2020benchmarking}, to ensure a fair comparison. We reported summary statistics (mean, and the standard deviation) of the performance metrics over four runs with different seeds of initialization. We used the same learning rate decay strategy as in \cite{dwivedi2020benchmarking} to train the models using the Adam optimizer~\cite{kingma2017adam}. More details for the implementation of the experiments are provided in the GitHub repository.

\subsection{GNN layers and network architecture}
\label{subsec:gnns}
We considered two \emph{isotropic} and two \emph{anisotropic} MP GNNs as baselines. We modified these SOTA models by applying the proposed \emph{bilateral}-MP algorithm and compared prediction performance with that of the original models. All GNNs consist of $L$-MP layers, and details of each SOTA GNN are reported in \autoref{appendix:details_on_gnn}. To apply the \emph{bilateral}-MP layer to the SOTA GNNs, we proposed three variations of network configuration: \emph{boosting}, \emph{interleaved}, and \emph{dense}. The \emph{boosting} configuration only adopts a \emph{bilateral}-MP layer at the initial part, while the other layers are unchanged. Specifically, only the second layer is modified to ensure better conditions for the soft cluster assignment networks in the \emph{bilateral}-MP layer. The \emph{interleaved} configuration alternates between ordinary and \emph{bilateral}-MP layers, and the \emph{dense} configuration consists only of \emph{bilateral}-MP layers. Heuristically, we reported the experimental results using only the \emph{boosting} configuration. The results and details for the other configurations can be found in \autoref{appendix:config}.We used the same setting as in \cite{dwivedi2020benchmarking} for any possible other basic building blocks, such as batch normalization, residual connection, and dropout. The hyperparameters for the number of layers and dimensions of the hidden and output representations for each layer are defined accordingly to match the parameter budgets proposed in \cite{dwivedi2020benchmarking}: 100k or 500k parameters. Finally, the task-specific networks which were mentioned in \autoref{subsec:mpgnns} are applied to the output of the stacked GNN layers, and other details for the implementation can be found in the GitHub repository.

\subsection{Benchmarking results}
\label{subsec:results1}
\paragraph{Graph-level predictions.} As shown in \autoref{tab:graphresult1}, the proposed \emph{bilateral}-MP exhibits consistently high performance across datasets and MP GNNs. All models with the \emph{bilateral}-MP framework outperform the baseline MP GNNs for the graph-level regression task on the ZINC dataset. Among the MP GNNs, the gated GCN with the \emph{bilateral}-MP (\emph{bi-gated} GCN) yields superior performance to that of other GNNs, including its baseline network (\emph{vanilla} gated GCN). Similarly, the best results in the graph classification task were yielded by the \emph{bi-gated} GCN. Results from the other baseline MP GNNs (GraphSAGE, GAT, and GCN) indicate that the proposed scheme boosts prediction performance, except in the cases of GAT on SMNIST and isotropic models on CIFAR10.

\paragraph{Node- and edge-level predictions.} \autoref{tab:graphresult2} reports the results for both the edge prediction and node classification tasks with the TSP and CLUSTER datasets. In the case of TSP, all GNNs using the proposed \emph{bilateral}-MP achieved higher performance than their respective baseline models. Our \emph{bilateral}-MP framework injects global-level information (the node class) in message aggregation to prevent over-localization. Thus, these results were expected because TSP graphs require reasoning about both local- (neighborhood nodes) and global-level (node class information) graph structures. On the CLUSTER dataset, the proposed framework adapts the GNNs to better represent the graphs that are highly modularized using the SBM. As a result, the experiments on CLUSTER also show that the performance of \emph{bilateral}-MP GNNs is consistently better than that of the baseline GNNs. The \emph{bi-gated} GCN achieves the best performance among the models.

\begin{table}
    \caption{Benchmarking results on four SOTA MP GNNs across datasets of graph-level prediction tasks. $L$ and $\text{Params}$ denote the number of layers and network parameters, respectively. We reported the statistics for mean absolute error (MAE), and prediction accuracy (ACC). $\text{Epochs}$ refers number of epochs for convergence. \textbf{Bold}: there are performance improvements with our \emph{bilateral}-MP scheme. \textcolor{red}{\textbf{Red}}: Best performance.}
    \resizebox{\textwidth}{!}{%
    \begin{tabular}{rrcccc|rrcccc}
    \toprule
    \multicolumn{12}{c}{ZINC}                   \\
    \cmidrule(r){1-12}
    Model & L & Params & Test MAE & Train MAE & Epochs & Model & L & Params & Test MAE & Train MAE & Epochs \\
    \midrule
    GCN  & 4 & 103077 & $0.459 \pm 0.006$ & $0.343 \pm 0.011$ & 196.25 & \emph{bi}-GCN & 4 & 125651 & \boldsymbol{$0.394 \pm 0.006$} & $0.300 \pm 0.005$ & 211.25     \\
    GraphSAGE  & 4 & 94977 & $0.468 \pm 0.003$ & $0.251 \pm 0.004$ & 147.25 & \emph{bi}-GraphSAGE & 4 & 115389 & \boldsymbol{$0.305 \pm 0.019$} & $0.152 \pm 0.022$ & 167.25     \\
    GAT  & 4 & 102385 & $0.475 \pm 0.007$ & $0.317 \pm 0.006$ & 137.5 & \emph{bi}-GAT & 4 & 105256 & \boldsymbol{$0.410 \pm 0.017$} & $0.247 \pm 0.058$ & 198.5     \\
    GatedGCN  & 4 & 105875 & $0.375 \pm 0.003$ & $0.236 \pm 0.007$ & 194.75 & \emph{bi}-GatedGCN & 4 & 111574 & \boldsymbol{$0.310 \pm 0.038$} & $0.186 \pm 0.060$ & 208.5     \\
    \hline
    GCN  & 16 & 505079 & $0.367 \pm 0.011$ & $0.128 \pm 0.019$ & 197 & \emph{bi}-GCN & 16 & 536482 & \boldsymbol{$0.276 \pm 0.007$} & $0.118 \pm 0.010$ & 161     \\
    GraphSAGE  & 16 & 505341 & $0.398 \pm 0.002$ & $0.081 \pm 0.009$ & 145.5 & \emph{bi}-GraphSAGE & 16 & 516651 & \boldsymbol{$0.245 \pm 0.009$} & $0.042 \pm 0.005$ & 157.5     \\
    GAT  & 16 & 531345 & $0.384 \pm 0.007$ & $0.067 \pm 0.004$ & 144 & \emph{bi}-GAT & 16 & 535536 & \boldsymbol{$0.277 \pm 0.012$} & $0.036 \pm 0.008$ & 148     \\
    GatedGCN  & 16 & 505011 & $0.214 \pm 0.013$ & $0.067 \pm 0.019$ & 185 & \emph{bi}-GatedGCN & 16 & 511974 & \color{red}{\boldsymbol{$0.166 \pm 0.009$}} & $0.049 \pm 0.015$ & 212.75     \\
    \hline
    \hline
    \multicolumn{12}{c}{SMNIST}                   \\
    \cmidrule(r){1-12}
    Model & L & Params & Test ACC & Train ACC & Epochs & Model & L & Params & Test ACC & Train ACC & Epochs \\
    \hline
    GCN  & 4 & 110807 & $90.705 \pm 0.218$ & $97.196 \pm 0.223$ & 127.5 & \emph{bi}-GCN & 4 & 104217 & \boldsymbol{$90.805 \pm 0.299$} & $98.210 \pm 0.411$ & 124.75     \\
    GraphSAGE  & 4 & 114169 & $97.312 \pm 0.097$ & $100.00 \pm 0.000$ & 98.25 & \emph{bi}-GraphSAGE & 4 & 110400 & \boldsymbol{$97.438 \pm 0.155$} & $100.00 \pm 0.000$ & 98     \\
    GAT  & 4 & 114507 & $95.535 \pm 0.205$ & $99.994 \pm 0.008$ & 104.75 & \emph{bi}-GAT & 4 & 104337 & $95.363 \pm 0.199$ & $100.00 \pm 0.000$ & 99.25     \\
    GatedGCN  & 4 & 125815 & $97.340 \pm 0.143$ & $100.00 \pm 0.000$ & 96.25 & \emph{bi}-GatedGCN & 4 & 101365 & \color{red}{\boldsymbol{$97.515 \pm 0.085$}} & $100.00 \pm 0.000$ & 110.25     \\
    \hline
    \hline
    \multicolumn{12}{c}{CIFAR10}                   \\
    \cmidrule(r){1-12}
    Model & L & Params & Test ACC & Train ACC & Epochs & Model & L & Params & Test ACC & Train ACC & Epochs \\
    \hline
    GCN  & 4 & 101657 & $55.710 \pm 0.381$ & $69.523 \pm 1.948$ & 142.5 & \emph{bi}-GCN & 4 & 125564 & $54.450 \pm 0.137$ & $69.455 \pm 1.975$ & 153.25     \\
    GraphSAGE  & 4 & 104517 & $65.767 \pm 0.308$ & $99.719 \pm 0.062$ & 93.5 & \emph{bi}-GraphSAGE & 4 & 114312 & $64.863 \pm 0.445$ & $99.878 \pm 0.083$ & 94     \\
    GAT  & 4 & 110704 & $64.223 \pm 0.455$ & $89.114 \pm 0.499$ & 103.75 & \emph{bi}-GAT & 4 & 114311 & \boldsymbol{$64.275 \pm 0.458$} & $87.670 \pm 0.627$ & 104.5     \\
    GatedGCN  & 4 & 104357 & $67.312 \pm 0.311$ & $94.553 \pm 1.018$ & 97 & \emph{bi}-GatedGCN & 4 & 110632 & \color{red}{\boldsymbol{$67.850 \pm 0.522$}} & $93.908 \pm 0.767$ & 103.75     \\
    \bottomrule
    \end{tabular}%
    }
    \label{tab:graphresult1}
\end{table}

\begin{table}
    \caption{Benchmarking results on four SOTA MP GNNs across datasets of edge- and node-level prediction tasks. We report the statistics for F1 score (F1), and prediction accuracy (ACC). \textbf{Bold}: there are performance improvements with our bilateral MP scheme. \textcolor{red}{\textbf{Red}}: Best performance.}
    \resizebox{\textwidth}{!}{%
    \begin{tabular}{rrcccc|rrcccc}
    \toprule
    \multicolumn{12}{c}{TSP}                   \\
    \cmidrule(r){1-12}
    Model & L & Params & Test F1 & Train F1 & Epochs & Model & L & Params & Test F1 & Train F1 & Epochs \\
    \midrule
    GCN  & 4 & 95702 & $0.630 \pm 0.001$ & $0.631 \pm 0.001$ & 261 & \emph{bi}-GCN & 4 & 118496 & \boldsymbol{$0.642 \pm 0.001$} & $0.644 \pm 0.001$ & 199     \\
    GraphSAGE  & 4 & 99263 & $0.665 \pm 0.003$ & $0.669 \pm 0.003$ & 266 & \emph{bi}-GraphSAGE & 4 & 131861 & \boldsymbol{$0.693 \pm 0.016$} & $0.696 \pm 0.015$ & 203.75     \\
    GAT  & 4 & 96182 & $0.673 \pm 0.002$ & $0.671 \pm 0.002$ & 328.25 & \emph{bi}-GAT & 4 & 115609 & \boldsymbol{$0.675 \pm 0.002$} & $0.677 \pm 0.002$ & 259.5     \\
    GatedGCN  & 4 & 97858 & $0.808 \pm 0.003$ & $0.811 \pm 0.003$ & 197 & \emph{bi}-GatedGCN & 4 & 125832 & \color{red}{\boldsymbol{$0.812 \pm 0.004$}} & $0.817 \pm 0.004$ & 235.25     \\
    \hline
    \hline
    \multicolumn{12}{c}{CLUSTER}                   \\
    \cmidrule(r){1-12}
    Model & L & Params & Test ACC & Train ACC & Epochs & Model & L & Params & Test ACC & Train ACC & Epochs \\
    \hline
    GCN  & 16 & 501687 & $68.498 \pm 0.976$ & $71.729 \pm 2.212$ & 79.75 & \emph{bi}-GCN & 16 & 505149 & \boldsymbol{$71.199 \pm 0.882$} & $75.196 \pm 1.640$ & 79.5     \\
    GraphSAGE  & 16 & 503350 & $63.844 \pm 0.110$ & $86.710 \pm 0.167$ & 57.75 & \emph{bi}-GraphSAGE & 16 & 490569 & \boldsymbol{$64.088 \pm 0.182$} & $86.815 \pm 0.343$ & 56.5     \\
    GAT  & 16 & 527824 & $70.587 \pm 0.447$ & $76.074 \pm 1.362$ & 73.5 & \emph{bi}-GAT & 16 & 445438 & \boldsymbol{$71.113 \pm 0.869$} & $74.834 \pm 2.737$ & 82.5     \\
    GatedGCN  & 16 & 504253 & $76.082 \pm 0.196$ & $88.919 \pm 0.720$ & 57.75 & \emph{bi}-GatedGCN & 16 & 516211 & \color{red}{\boldsymbol{$76.896 \pm 0.213$}} & $87.954 \pm 0.365$ & 56.5     \\
    \bottomrule
    \end{tabular}%
    }
    \label{tab:graphresult2}
\end{table}

\paragraph{Small-size benchmark datasets.} In addition to these medium-size benchmarking results, we also performed graph classification tasks using three other small-sized datasets: DD~\cite{dd}, PROTEINS~\cite{borgwardt2005protein}, and ENZYMES~\cite{enzymes}. We reported the details and results of the experiments on small-sized datasets in \autoref{appendix:small}.

\subsection{Quantitative feature analysis}
\label{subsec:results2}

\captionsetup[figure]{font=small,skip=0pt}
\begin{wrapfigure}{r}{0.60\textwidth}
    \vspace{-22pt}
    \includegraphics[width=0.60\textwidth]{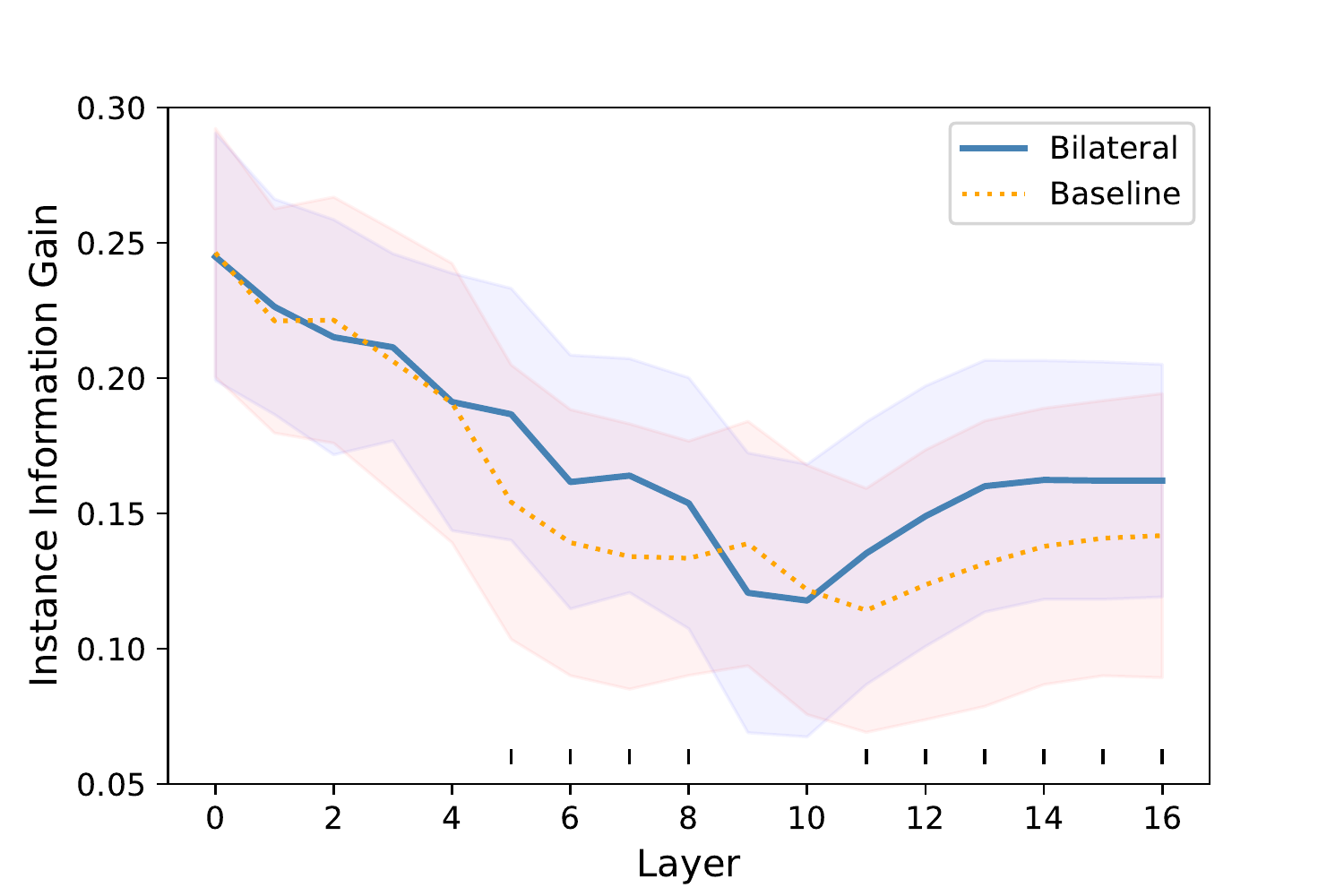}
    \caption{Comparison of the instance information gain between \textcolor{blue}{\emph{bi}-GCN} and \textcolor{orange}{\emph{vanilla} GCN} across the layers. The solid (and dashed) lines and shaded region denote mean and standard deviation for the 1$k$ test graphs of CLUSTER, respectively. Short vertical bars indicate a statistically significant difference in the two-sampled T-tests between the baseline and proposed models after Bonferroni multiple-comparison correction.}
    \label{fig:Figure2}
\end{wrapfigure}
\captionsetup[figure]{font=normal,skip=0pt}

In \autoref{subsec:results1}, we demonstrated the robustness and effectiveness of our proposed framework on graph machine-learning tasks. In this section, we performed an additional analysis to understand how the \emph{bilateral}-MP prevents over-smoothing. We quantitatively measured over-smoothing to investigate how it changes along the layers. To eliminate any undesired effect, we adopted the simplest model and its variant: GCN with and without the \emph{boosting} \emph{bilateral}-MP layer (\emph{bi}-GCN). With the same setting as the above benchmarking experiments, these GNNs were trained and subsequently used to infer the test graphs using the CLUSTER dataset. Given a test graph $G$, we obtained the instance information gain $\bm{I}_{G}^{l}$ based on the information theory at each layer $l$ \cite{dgn}.

\begin{equation}
   \bm{I}_{G}^{l} = \sum_{v} \bm{P}_{\bm{X},\bm{H^l}}(\bm{x}_v,\bm{h}_v) \log {\frac {\bm{P}_{\bm{X},\bm{H^l}}(\bm{x}_v,\bm{h}_v)} {\bm{P}_{\bm{X}}(\bm{x}_v) \bm{P}_{\bm{H^l}}(\bm{h}_v) }}
   \label{eq:iig}
\end{equation}

where $\bm{P}_{\bm{X}}$, and $\bm{P}_{\bm{H^l}}$ are the probability distributions of the input feature, and the hidden representation at layer $l$, respectively, both defined under the Gaussian assumption. $\bm{P}_{\bm{X},\bm{H^l}}$ denotes joint distribution. The instance information gain is a measurement of the mutual information between the input feature and hidden representation. Thus, the values of $\bm{I}_{G}^{l}$ tend to decrease with the intensification of over-smoothing, as node embeddings lose their input information by averaging neighbor information. As shown in Figure 3, the information gains of the proposed model are significantly higher than those of the baseline model along the layers. This result suggests that our \emph{bilateral}-MP framework can prevent the over-smoothing issue.
\section{Conclusion}

In this study, we introduced a new \emph{bilateral}-MP scheme to minimize over-smoothing. To prevent over-mixing of node representations across different classes, the proposed model regulates the aggregating layer by utilizing the global community structure with the class information of nodes. Our model boosts the performance of existing GNNs by preventing over-smoothing. We expect to contribute to the community by providing a simple insight and sanity-check for novel MP GNNs in the future.

\newpage

\begin{ack}
This work was supported by Institute of Information \& communications Technology Planning \& Evaluation (IITP) grant funded by the Korea government(MSIT) (No.2020$-$0$-$01373, Artificial Intelligence Graduate School Program(Hanyang University))
\end{ack}

{
\small
\bibliographystyle{plainnat} 
\bibliography{references}
}

%
%

\clearpage


\makeatletter
  \setcounter{table}{0}
  \renewcommand{\thetable}{S\arabic{table}}%
  \setcounter{figure}{0}
  \renewcommand{\thefigure}{S\arabic{figure}}%
  \setcounter{equation}{0}
  \renewcommand\theequation{S\arabic{equation}}
  \renewcommand{\bibnumfmt}[1]{[S#1]}

  \newcommand{\suptitle}{\titl \\ --- Supplementary material ---}
  \renewcommand{\@title}{\suptitle}
  \renewcommand{\@author}{}

  \par
  \begingroup
    \renewcommand{\thefootnote}{\fnsymbol{footnote}}
    \renewcommand{\@makefnmark}{\hbox to \z@{$^{\@thefnmark}$\hss}}
    \renewcommand{\@makefntext}[1]{%
      \parindent 1em\noindent
      \hbox to 1.8em{\hss $\m@th ^{\@thefnmark}$}#1
    }
    \thispagestyle{empty}
    \@maketitle
  \endgroup
  \let\maketitle\relax
  \let\thanks\relax
\makeatother

\appendix

\vspace{-2cm}
\section{Details of the bilateral-GNN layer}\label{appendix:details_on_gnn}

In this section, we present details for the SOTA MP layers and their bilateral variants, which were used for the benchmarking experiments in \autoref{subsec:results1}. The MP GNNs, consisting of $L$-layers that embed the node representations, can consider multi-hop neighborhoods~\cite{dgn}. 

\paragraph{Graph Convolutional Networks (GCN).}
GCN layers can be implemented as a simple MP using some approximations and simplifications of the Chebyshev polynomials~\cite{kipf2017semisupervised}:

\begin{equation}
    \bm{h}_{v}^{l+1}=\text{ReLU}(\bm{U}^{l} {\frac {1}{\text{deg}_{v}}} \sum_{u \in \mathcal{N}_{v}}\bm{h}_{u}^{l} )
    \label{eq:gcn}
\end{equation}

where $\text{deg}_{v}$ denotes the degree of node $v$, and $\bm{U}^{l} \in \mathbb{R}^{D \times D}$ is a trainable parameter. We applied modular gradient weighting in \autoref{subsec:modulargradient} to the aggregating function of \autoref{eq:gcn} to define the \emph{bi}-GCN layer:

\begin{equation}
    \bm{h}_{v}^{l+1}=\text{ReLU}(\bm{U}^{l} {\frac {\hat{\beta}_{u,v}^{l}}{\text{deg}_{v}}} \sum_{u \in \mathcal{N}_{v}}\bm{h}_{u}^{l} )
    \label{eq:bigcn}
\end{equation}

\paragraph{GraphSAGE.}
Following \cite{dwivedi2020benchmarking}, we use the GraphSAGE layer utilizing a max aggregator~\cite{hamilton2018inductive}:

\begin{equation}
    \bm{h}_{v}^{l+1}=\text{ReLU}( \bm{U}^{l} \text{Concat}( \bm{h}_{v}^{l} , \text{Max}_{u \in \mathcal{N}_{v}}\text{ReLU}( \bm{V}^{l} \bm{h}_{u}^{l} ) )
    \label{eq:graphsage}
\end{equation}

where $\text{Max}(\cdot)$ is the maxpooling operator, and $\bm{U}^{l} \in \mathbb{R}^{D \times D}$, and $\bm{V}^{l} \in \mathbb{R}^{D \times D}$ are trainable parameters. We applied modular gradient weighting in \autoref{subsec:modulargradient} to the aggregating function of \autoref{eq:graphsage} to define the \emph{bi}-GraphSAGE layer:

\begin{equation}
    \bm{h}_{v}^{l+1}=\text{ReLU}( \bm{U}^{l} \text{Concat}( \bm{h}_{v}^{l} , \text{Max}_{u \in \mathcal{N}_{v}}\hat{\beta}_{u,v}^{l}\text{ReLU}( \bm{V}^{l} \bm{h}_{u}^{l} ) )
    \label{eq:bigraphsage}
\end{equation}

\paragraph{Graph Attention Networks (GAT).}
GAT layers embed representation of a center node by applying multi-head self-attention strategy with representations of neighborhood nodes~\cite{gat}. The pairwise self-attention scores $e_{u,v}^{k,l}$ at layer $l$ and head $k$ were normalized and used to perform \emph{anisotropic} propagation of the neighborhood messages:

\begin{equation}
    \bm{h}_{v}^{l+1}=\text{Concat}_{k=1}^{K}( \text{ELU}( \sum_{u \in \mathcal{N}_{v}}e_{u,v}^{k,l}\bm{U}^{k,l} \bm{h}_{u}^{l} ) )
    \label{eq:gat1}
\end{equation}
\begin{equation}
    e_{u,v}^{k,l} = {\frac {\text{exp}(\hat{e}_{u,v}^{k,l})} {\sum_{\Dot{u} \in \mathcal{N}_{v}} {\text{exp}(\hat{e}_{\Dot{u},v}^{k,l})}}}
    \label{eq:gat2}
\end{equation}
\begin{equation}
    \hat{e}_{u,v}^{k,l} = \text{lReLU}( \bm{V}^{k,l} \text{Concat}( \bm{U}^{k,l}\bm{h}_{v}^{l} , \bm{U}^{k,l}\bm{h}_{u}^{l} ) )
    \label{eq:gat3}
\end{equation}

where $\text{ELU}$ and $\text{lReLU}$ denote the exponential linear unit (ELU)~\cite{elu}, and leaky ReLU nonlinearity~\cite{lrelu} functions, respectively. We applied modular gradient weighting in \autoref{subsec:modulargradient} to the aggregating function of \autoref{eq:gat1} to define the \emph{bi}-GAT layer:

\begin{equation}
    \bm{h}_{v}^{l+1}=\text{Concat}_{k=1}^{K}( \text{ELU}( \sum_{u \in \mathcal{N}_{v}}\hat{\beta}_{u,v}^{l}e_{u,v}^{k,l}\bm{U}^{k,l} \bm{h}_{u}^{l} ) )
    \label{eq:bigat}
\end{equation}

\paragraph{Residual gated graph neural networks (gated GCN).}
The layer introduced in \cite{bresson2017residual} embeds representations of nodes $v$ ($\bm{h}_{v}^{l}$) and its edges ($\bm{e}_{u,v}^{l}, \forall u \in \mathcal{N}_{v}$) simultaneously with a residual mechanism. The edge embeddings $e$ can also be used as \emph{anisotropic} aggregating weights.

\begin{equation}
    \bm{h}_{v}^{l+1}=\bm{h}_{v}^{l}+\text{ReLU}( \text{BN}( \bm{U}^{l}\bm{h}_{v}^{l} + 
    \sum_{u \in \mathcal{N}_{v}} \bm{e}_{u,v}^{l} \otimes \bm{V}^{l}\bm{h}_{u}^{l} ) )
    \label{eq:gatedgcn1}
\end{equation}
\begin{equation}
    \bm{e}_{u,v}^{l} = {\frac {\sigma(\hat{\bm{e}}_{u,v}^{l})} {\sum_{\Dot{u} \in \mathcal{N}_{v}} {\sigma(\hat{\bm{e}}_{\Dot{u},v}^{l})} + \epsilon }}
    \label{eq:gatedgcn2}
\end{equation}
\begin{equation}
    \hat{\bm{e}}_{u,v}^{l}=\hat{\bm{e}}_{u,v}^{l-1}+\text{ReLU}( \text{BN}( \bm{A}^{l}\bm{h}_{u}^{l-1} + \bm{B}^{l}\bm{h}_{v}^{l-1} + \bm{C}^{l}\hat{\bm{e}}_{u,v}^{l-1} ) )
    \label{eq:gatedgcn3}
\end{equation}

where $ \text{BN}(\cdot)$ is batch normalization~\cite{ioffe2015batch} and $\otimes$ is element-wise multiplication. $\bm{U}^{l} \in \mathbb{R}^{D \times D}$, $\bm{V}^{l} \in \mathbb{R}^{D \times D}$, $\bm{A}^{l} \in \mathbb{R}^{D \times D}$, $\bm{B}^{l} \in \mathbb{R}^{D \times D}$, and $\bm{C}^{l} \in \mathbb{R}^{D \times D}$ are trainable parameters. We applied modular gradient weighting in \autoref{subsec:modulargradient} to the aggregating function of \autoref{eq:gatedgcn1} to define the \emph{bi}-GatedGCN layer:

\begin{equation}
    \bm{h}_{v}^{l+1}=\bm{h}_{v}^{l}+\text{ReLU}( \text{BN}( \bm{U}^{l}\bm{h}_{v}^{l} + 
    \sum_{u \in \mathcal{N}_{v}} \bm{e}_{u,v}^{l} \otimes \hat{\beta}_{u,v}^{l}\bm{V}^{l}\bm{h}_{u}^{l} ) )
    \label{eq:bigatedgcn}
\end{equation}
\section{Results for the other configuration of the bilateral-GNN Network architecture}\label{appendix:config}

\begin{figure}[t!]
    \centering
    \includegraphics[width=\textwidth]{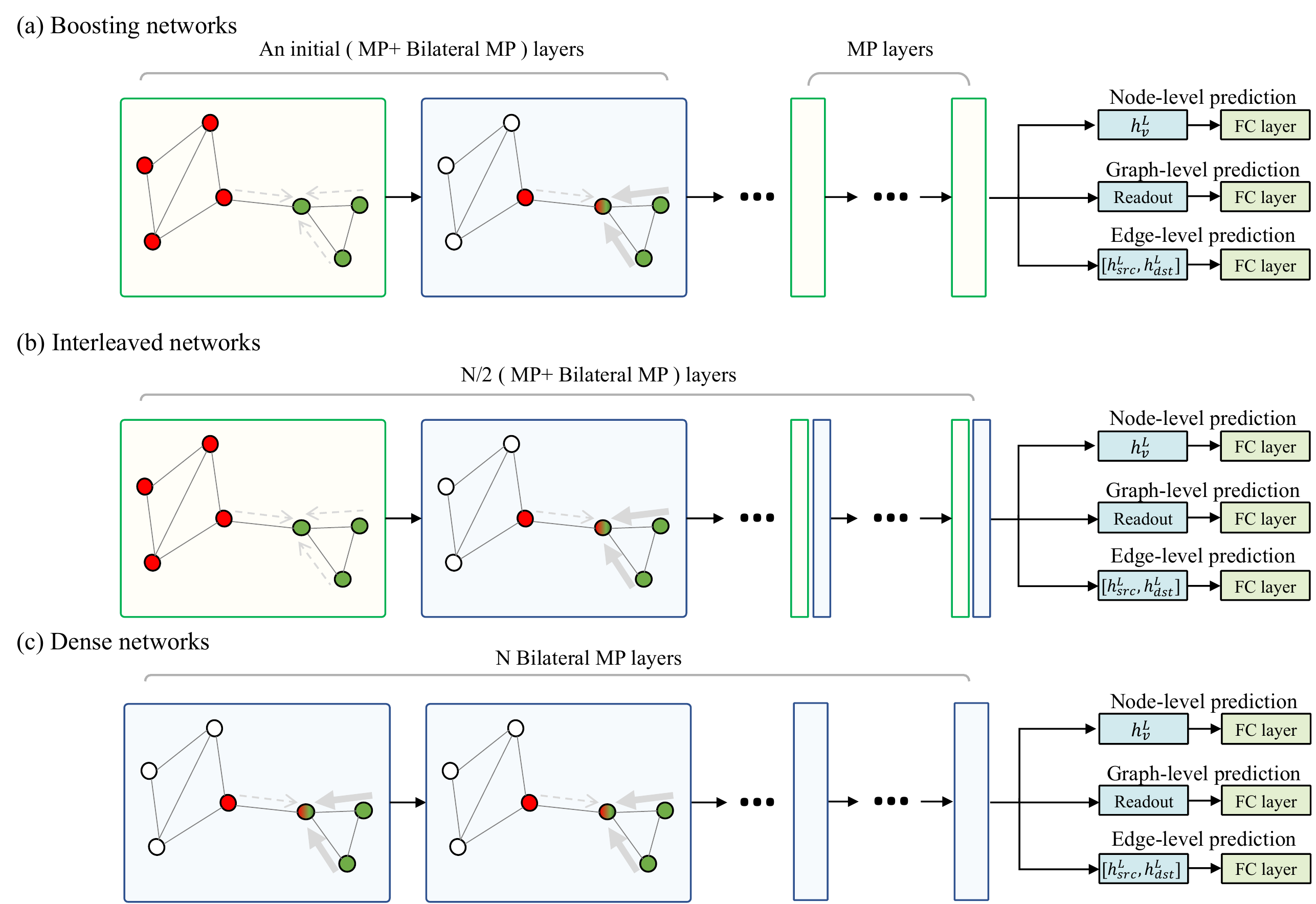}
    \caption{Three possible configurations of GNNs using the proposed bilateral MP layer(s).}
    \label{fig:FigureS1}
\end{figure}

\autoref{fig:FigureS1} illustrates three configurations of network architecture using the \emph{bilateral}-MP layer(s). A comparison of prediction performance across the configurations was performed using the gated GCN MP layer and ZINC dataset. As shown in \autoref{tab:tables1}, the \emph{boosting} configuration achieved the best performance, while the other configurations were suboptimal.

\begin{table}
    \caption{Benchmarking results on gated GCN on the ZINC dataset across three network configurations. \textbf{Bold}: there are performance improvements with our \emph{bilateral}-MP scheme. \textcolor{red}{\textbf{Red}}: Best performance.}
    \resizebox{\textwidth}{!}{%
    \begin{tabular}{r|rcccc|rcccc}
    \toprule
    Model & L & Params & Test MAE & Train MAE & Epochs & L & Params & Test MAE & Train MAE & Epochs \\
    \midrule
    GatedGCN  & 4 & 105875 & $0.375 \pm 0.003$ & $0.236 \pm 0.007$ & 194.75 & 16 & 505011 & $0.214 \pm 0.013$ & $0.067 \pm 0.019$ & 185     \\
    \emph{boosting}-GatedGCN  & 4 & 111574 & \boldsymbol{$0.310 \pm 0.038$} & $0.186 \pm 0.060$ & 208.5 & 16 & 511974 & \color{red}{\boldsymbol{$0.166 \pm 0.009$}} & $0.049 \pm 0.015$ & 212.75     \\
    \emph{interleaved}-GatedGCN  & 4 & 117273 & \boldsymbol{$0.372 \pm 0.011$} & $0.220 \pm 0.028$ & 180.75 & 16 & 560715 & \boldsymbol{$0.213 \pm 0.012$} & $0.073 \pm 0.005$ & 192.75     \\
    \emph{dense}-GatedGCN  & 4 & 128671 & \boldsymbol{$0.369 \pm 0.025$} & $0.204 \pm 0.031$ & 205 & 16 & 616419 & \boldsymbol{$0.210 \pm 0.013$} & $0.064 \pm 0.012$ & 189.75     \\
    \bottomrule
    \end{tabular}%
    }
    \label{tab:tables1}
\end{table}
\section{Benchmarking results for small-size dataset}\label{appendix:small}

Additionally, we demonstrated that the proposed \emph{bilateral}-MP framework works well on several popular small datasets. We performed experiments on three TU graph classification datasets~\cite{morris2020tudataset} with the same settings (data split, learning rate strategy, and batch size) as in \cite{dwivedi2020benchmarking}. A detailed description of the data statistics is provided in \autoref{tab:dataset2}. 

\begin{table}[hbt!]
    \centering
    \caption{Statistics on three small benchmark datasets of TU. Columns 3 and 4 denote mean statistics for the number of nodes and edges, respectively. Columns 5 and 6 summarize the node and edge features (and dimension), respectively.}
    \resizebox{0.7\textwidth}{!}{
    \begin{tabular}{lcrrcc}
    \toprule
    \textbf{Dataset} & \textbf{\#Graphs} & \textbf{\#Nodes} & \textbf{\#Edges} &  \textbf{Node feat (dim)} &
\textbf{Edge feat (dim)} \\
    \midrule
    DD & 1178 & 284.32  & 715.66 &  Node labels (89) & $-$    \\
    ENZYMES & 600 & 32.63  & 62.14 &  Node attributes (18) & $-$    \\
    PROTEINS & 1113 & 39.06  & 72.82 &  Node attributes (29) & $-$    \\
    \bottomrule
    \end{tabular}%
    }
    \label{tab:dataset2}
\end{table}

As shown in \autoref{tab:smallresults}, the \emph{bi}-GatedGCN and \emph{bi}-GraphSAGE were consistently higher performing than the original MPs. In contrast, \emph{bi}-GAT could not outperform the baseline GAT model. For the GCN, only the \emph{bi}-GCN in the DD dataset exhibited a boost in prediction performance.

\begin{table}[hbt!]
    \caption{Benchmarking results on four SOTA MP GNNs across datasets of small graph-level classification tasks. \textbf{Bold}: there are performance improvements with our \emph{bilateral}-MP scheme. \textcolor{red}{\textbf{Red}}: Best performance.}
    \resizebox{\textwidth}{!}{%
    \begin{tabular}{rrcccc|rrcccc}
    \toprule
    \multicolumn{12}{c}{DD}                   \\
    \cmidrule(r){1-12}
    Model & L & Params & Test ACC & Train ACC & Epochs & Model & L & Params & Test ACC & Train ACC & Epochs \\
    \midrule
    GCN  & 4 & 102293 & $72.758 \pm 4.083$ & $100.00 \pm 0.000$ & 266.7 & \emph{bi}-GCN & 4 & 117271 & \boldsymbol{$73.517 \pm 4.321$} & $100.00 \pm 0.000$ & 280     \\
    GraphSAGE  & 4 & 102577 & $73.433 \pm 3.429$ & $100.00 \pm 0.000$ & 267.2 & \emph{bi}-GraphSAGE & 4 & 113435 & \boldsymbol{$74.193 \pm 2.482$} & $100.00 \pm 0.000$ & 267.2     \\
    GAT  & 4 & 100132 & $75.900 \pm 3.824$ & $95.851 \pm 2.575$ & 201.3 & \emph{bi}-GAT & 4 & 117767 & $74.372 \pm 3.716$ & $99.894 \pm 0.134$ & 109.8     \\
    GatedGCN  & 4 & 104165 & $72.918 \pm 2.090$ & $82.796 \pm 2.242$ & 300.7 & \emph{bi}-GatedGCN & 4 & 105227 & \boldsymbol{$73.862 \pm 3.069$} & $94.014 \pm 2.088$ & 237.2     \\
    \hline
    \hline
    \multicolumn{12}{c}{ENZYMES}                   \\
    \cmidrule(r){1-12}
    Model & L & Params & Test ACC & Train ACC & Epochs & Model & L & Params & Test ACC & Train ACC & Epochs \\
    \hline
    GCN  & 4 & 103407 & $65.833 \pm 4.610$ & $97.688 \pm 3.064$ & 343 & \emph{bi}-GCN & 4 & 113641 & $60.333 \pm 6.316$ & $87.500 \pm 8.079$ & 338.4     \\
    GraphSAGE  & 4 & 105595 & $65.000 \pm 4.944$ & $100.00 \pm 0.000$ & 294.2 & \emph{bi}-GraphSAGE & 4 & 115033 & \boldsymbol{$68.333 \pm 2.687$} & $100.00 \pm 0.000$ & 303.1     \\
    GAT  & 4 & 101274 & $68.500 \pm 5.241$ & $100.00 \pm 0.000$ & 299.3 & \emph{bi}-GAT & 4 & 104420 & $67.667 \pm 3.350$ & $99.771 \pm 0.286$ & 299.5     \\
    GatedGCN  & 4 & 103409 & $65.667 \pm 4.899$ & $99.979 \pm 0.062$ & 316.8 & \emph{bi}-GatedGCN & 4 & 106237 & \boldsymbol{$68.167 \pm 5.747$} & $100.00 \pm 0.000$ & 322.2     \\
    \hline
    \hline
    \multicolumn{12}{c}{PROTEINS}                   \\
    \cmidrule(r){1-12}
    Model & L & Params & Test ACC & Train ACC & Epochs & Model & L & Params & Test ACC & Train ACC & Epochs \\
    \hline
    GCN  & 4 & 104865 & $76.098 \pm 2.406$ & $81.387 \pm 2.451$ & 350.9 & \emph{bi}-GCN & 4 & 116005 & $76.001 \pm 3.545$ & $81.713 \pm 2.166$ & 339     \\
    GraphSAGE  & 4 & 101928 & $75.289 \pm 2.419$ & $85.827 \pm 0.839$ & 245.4 & \emph{bi}-GraphSAGE & 4 & 114199 & \color{red}{\boldsymbol{$76.996 \pm 2.933$}} & $87.704 \pm 1.522$ & 242.6     \\
    GAT  & 4 & 102710 & $76.277 \pm 2.410$ & $83.186 \pm 2.000$ & 344.6 & \emph{bi}-GAT & 4 & 110556 & $74.751 \pm 1.834$ & $83.871 \pm 1.738$ & 192.3     \\
    GatedGCN  & 4 & 104855 & $76.363 \pm 2.904$ & $79.431 \pm 0.695$ & 293.8 & \emph{bi}-GatedGCN & 4 & 99830 & \boldsymbol{$76.453 \pm 3.109$} & $79.858 \pm 0.449$ & 337     \\
    \bottomrule
    \end{tabular}%
    }
    \label{tab:smallresults}
\end{table}

\end{document}